\documentclass{article}
\usepackage{microtype}          
\usepackage{graphicx}           
\usepackage{subcaption}         
\usepackage{booktabs}           
\usepackage{hyperref}           

\usepackage[accepted]{icml2026}

\usepackage{amsmath}
\usepackage{amssymb}
\usepackage{amsfonts}
\usepackage{mathtools}
\usepackage{amsthm}
\usepackage{bm}                 

\usepackage{algorithm}
\usepackage{algorithmic}

\usepackage[capitalize,noabbrev]{cleveref}

\theoremstyle{plain}

\theoremstyle{definition}

\theoremstyle{remark}

\icmltitlerunning{LaGO: Latent Action Guidance for Online Reinforcement Learning}

\begin{document}

\twocolumn[
  \icmltitle{LaGO: Latent Action Guidance for Online Reinforcement Learning}

  \icmlsetsymbol{equal}{*}

  \begin{icmlauthorlist}
    \icmlauthor{Kuan-Yen Liu}{equal,inst1}
    \icmlauthor{Ren-Jyun Huang}{equal,inst2}
    \icmlauthor{Ti-Rong Wu}{inst3}
  \end{icmlauthorlist}

    \icmlaffiliation{inst1}{Siebel School of Computing and Data Science, University of Illinois Urbana-Champaign, USA}
  \icmlaffiliation{inst2}{Department of Computer Science, National Yang Ming Chiao Tung University, Taiwan}
  \icmlaffiliation{inst3}{Institute of Information Science, Academia Sinica, Taiwan}
  \icmlcorrespondingauthor{Kuan-Yen Liu}{kuanyen2@illinois.edu}

  \icmlkeywords{%
    Reinforcement Learning,
    Large Language Models,
    World Models,
    Behavioral Priors,
    Continuous Control,
    ICML
  }

  \vskip 0.3in
]

\printAffiliationsAndNotice{\icmlEqualContribution}

%
%

\begin{abstract}

Large language models (LLMs) have shown strong potential for planning and sequential decision-making, but prior work often relies on using them as direct controllers, which requires precise action generation and can be unreliable in practice.
This paper proposes \textit{Latent Action Guidance for Online Reinforcement Learning (LaGO)}, a framework that uses a pretrained LLM as a latent action prior to softly guide online policy optimization, rather than treating the LLM as an explicit planner or controller.
Experiments on both a discrete-control benchmark, CLEVR-Robot, and a continuous-control benchmark, Meta-World, demonstrate that LaGO consistently improves both reward and success rate over Vanilla PPO.
In particular, LaGO increases the average success rate from 15.1\% to 27.2\% on CLEVR-Robot and from 2.7\% to 15.2\% on Meta-World.
Our analysis further shows that stronger pretrained LLMs provide more effective guidance, suggesting that LLM knowledge can improve planning and online decision-making.
\end{abstract}

\section{Introduction}
\label{sec:introduction}


Large language models (LLMs) have shown remarkable capabilities across a wide range of tasks \cite{openai2024gpt4technicalreport,touvron2023llamaopenefficientfoundation,Guo_2025,yang2025qwen3technicalreport}, suggesting that pretrained models acquire rich knowledge about environments, tasks, and possible behaviors.
Motivated by this, recent studies have explored the use of LLMs in reinforcement learning (RL) and sequential decision-making problems.
For example, some works use LLMs as direct controllers \cite{brohan2023rt2visionlanguageactionmodelstransfer,reed2022generalistagent,yao2023reactsynergizingreasoningacting}, where the model predicts actions from observations or multimodal inputs.
Other works use LLMs as high-level planners \cite{ahn2022icanisay,huang2022innermonologueembodiedreasoning,wang2023voyageropenendedembodiedagent,hu2026occubenchevaluatingaiagents}, where the model proposes subgoals or task plans and a separate low-level RL policy executes them.
Furthermore, LLMs have also been used as internal world model simulators \cite{hao2023reasoninglanguagemodelplanning,lin2024learningmodelworldlanguage,bruce2024geniegenerativeinteractiveenvironments,dataflowteam2026openworldlibunifiedcodebasedefinition,fang2025webevolverenhancingwebagent}, where the model generates imagined future trajectories to predict environment dynamics for planning.
These results highlight the potential of LLMs as a useful component for planning and RL.

Despite these promising results, existing approaches often require LLMs to provide sufficiently accurate outputs, such as actions, plans, or imagined trajectories, in order to be useful for RL control.
However, this requirement can be overly strong in practice, since the capabilities of LLMs may vary substantially across different models.
As a result, directly relying on an LLM as the primary controller may be unreliable, especially in RL problems that require precise execution and long-horizon reasoning.
In such settings, even small prediction errors may accumulate over time and lead to poor control performance.
Therefore, although LLMs may contain useful prior knowledge, using them for explicit control in RL remains challenging.

To address this challenge, this paper investigates whether LLMs can be used not as explicit controllers but as a source of soft guidance for RL.
Specifically, we propose \textit{Latent Action Guidance for Online Reinforcement Learning (LaGO)}, a two-stage framework for leveraging LLM knowledge in online RL.
In the first stage, LaGO uses expert demonstrations to fine-tune a pretrained LLM to learn a latent action guidance model.
Then, the learned model provides action guidance signals as inductive priors for policy learning during online RL training.
This substantially relaxes the accuracy requirement of using LLMs as direct controllers, since the LLM only needs to provide a coarse but informative guidance signal rather than exact decisions.

Experiments on both discrete control tasks on CLEVR-Robot and continuous control tasks in Meta-World demonstrate that LaGO consistently improves both reward and success rate over vanilla PPO. Specifically, it increases the average success rate from 15.1\% to 27.2\% in CLEVR-Robot, and from 2.7\% to 15.2\% in Meta-World.
The results demonstrate that LLMs can still help RL training even when they are not sufficiently accurate to serve as direct controllers.
Furthermore, our results show that the quality of the pretrained LLM models has a substantial impact on the effectiveness of the proposed method.
These findings suggest that stronger foundation LLMs are more likely to provide useful guidance signals for RL training in the future.
Overall, LaGO provides a simple and practical framework for leveraging pretrained LLM knowledge as latent action guidance for planning and RL.

\section{Related Work}
\label{sec:related_work}

\subsection{LLMs as Direct or Hierarchical Controllers}
\label{subsec:direct_hierarchical}

Recent progress in Vision-Language-Action (VLA) models \cite{brohan2023rt2visionlanguageactionmodelstransfer,driess2023palmeembodiedmultimodallanguage,kim2024openvlaopensourcevisionlanguageactionmodel} has demonstrated that large-scale pretraining foundation LLM models can be adapted to directly output control signals in embodied environments.
These methods demonstrate the potential of using foundation models as direct controllers, especially when rich visual and language inputs are available.
However, using LLMs as end-to-end controllers remains challenging in reinforcement learning settings, as it requires precise action execution, particularly in vectorized environments where observations and actions are not naturally represented in language or visual form.
An alternative approach is to adopt a hierarchical paradigm \cite{ahn2022icanisay,wang2023voyageropenendedembodiedagent,huang2022innermonologueembodiedreasoning}, where the LLM serves as a high-level planner that proposes textual subgoals or instructions, while a separate low-level policy handles execution.
This design avoids requiring the LLM to perform direct low-level control, but still relies on explicit textual outputs, and effectively integrating these outputs into RL policy learning remains a challenge.

\subsection{LLMs as World Models for Planning}
\label{subsec:llms_world_models}

LLMs have been shown to encode structured world knowledge about environments, including task dynamics, physical regularities, and action consequences, acquired during large-scale pretraining \cite{li2024emergentworldrepresentationsexploring,bubeck2023sparksartificialgeneralintelligence}.
Recent studies have further provided evidence that world-relevant information is preserved in LLM hidden representations.
For example, \citet{gurnee2024languagemodelsrepresentspace} demonstrate that spatial and temporal information can be linearly decoded from LLM hidden states, while \citet{jin2024emergentrepresentationsprogramsemantics} show that sequence models can encode future environment configurations in their latent state.
Motivated by this, several works have attempted to leverage such world knowledge from LLMs for planning and sequential decision-making.
\citet{lin2024learningmodelworldlanguage} incorporate language into a world model to improve future prediction and policy learning.
\citet{xiang2023languagemodelsmeetworld} improve language models by finetuning them on embodied experiences collected from simulators.



Despite these advances, these approaches still require LLMs to explicitly and accurately simulate valid future states or trajectories.
To address this, recent work such as KALM \cite{pang2024knowledgeableagentsofflinereinforcement} first fine-tunes an LLM on expert demonstrations, and then uses the fine-tuned model to generate additional rollouts under offline RL settings.
While effective, this approach does not directly extend to the online setting.
In contrast, our work investigates whether LLMs can also improve reinforcement learning in the online setting.

\subsection{LLMs as Priors for Reinforcement Learning}
\label{subsec:efficient_rl_priors}


In reinforcement learning, another natural idea is to introduce a prior to guide policy learning and exploration.
Recent work has begun to explore LLMs in this direction.
For example, \citet{yan2024efficientreinforcementlearninglarge} treat LLMs as the prior action distribution and incorporate them into both policy-based and value-based RL frameworks from a Bayesian inference perspective, demonstrating that fixed LLM priors can improve learning efficiency.
However, this and related approaches \cite{yao2023reactsynergizingreasoningacting,carta2026groundinglargelanguagemodels} are still limited to text-based or highly discretized action spaces.

In conclusion, compared to previous works, our work investigates a different way of leveraging LLMs in reinforcement learning.
Rather than using LLMs as direct controllers, explicit world model simulators, or text-based action priors, we study whether their knowledge can benefit online RL through latent representations rather than explicit text.
Instead of requiring the LLM to generate precise actions, plans, or trajectories during interaction, we use it as a source of latent guidance that guides policy learning without relying on explicit text.

\section{Method}
\label{sec:method}

\begin{figure*}[t]
    \centering
    \includegraphics[width=0.95\linewidth]{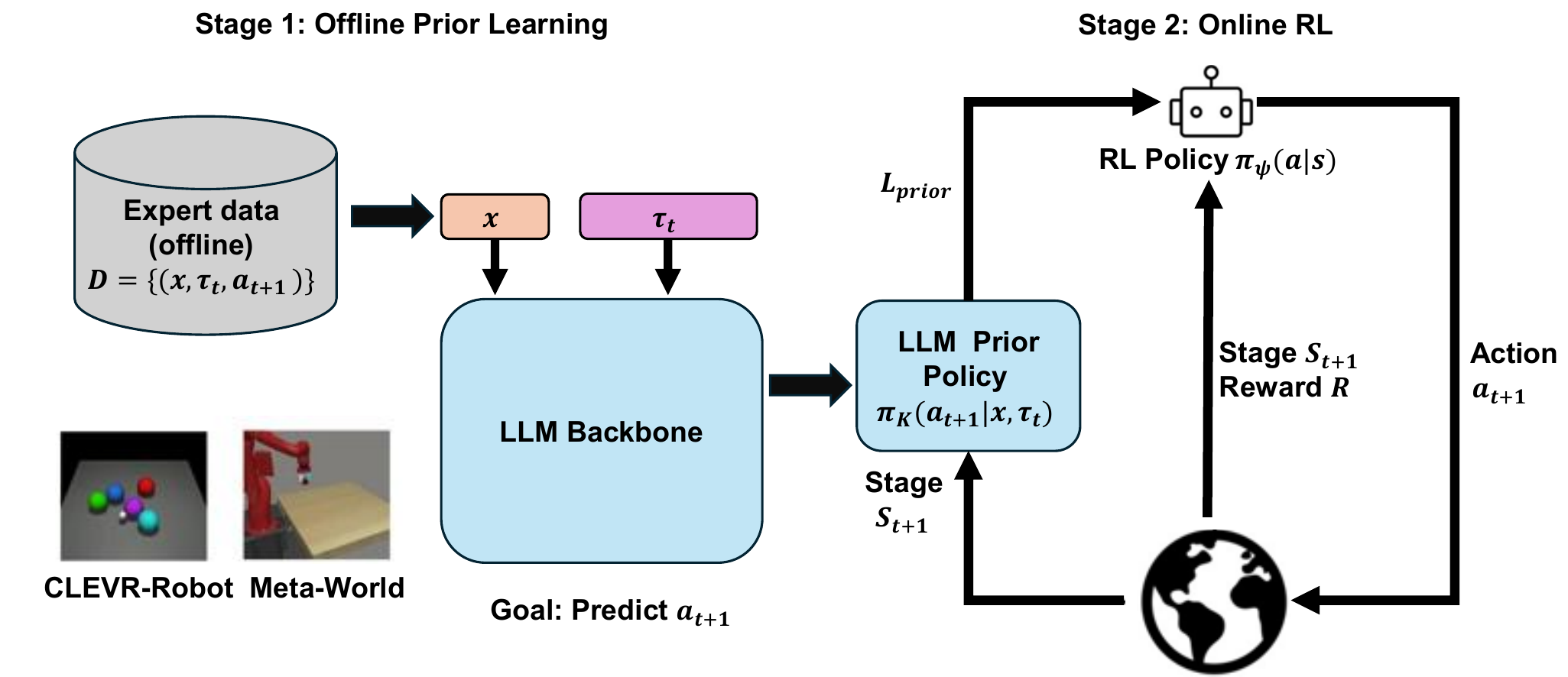}
    \caption{%
        Overview of the LaGO framework.
        Numeric environment states are injected into the frozen LLM latent space via learned projection layers, and the resulting action distribution serves as a KL-regularized prior for online RL policy optimization.
    }
    \label{fig:kalm_framework}
\end{figure*}

This paper presents \textit{Latent Action Guidance for Online Reinforcement Learning (LaGO)}, a framework that leverages pretrained LLM knowledge as a latent behavioral prior for online reinforcement learning.
As illustrated in Figure~\ref{fig:kalm_framework}, LaGO contains two stages.
First, LaGO trains a latent policy model from offline demonstrations that maps environment states to action distributions through a pretrained language model backbone.
Second, this model serves as a behavioral prior to guide online reinforcement learning.
We describe these two stages in detail below.

\subsection{Learning a Latent Policy Prior from Offline Demonstrations}
\label{subsec:lago_stage1}

Pretrained LLMs have been shown to encode rich knowledge about environments, tasks, and possible behaviors, suggesting that they can serve as a useful source of behavioral priors for reinforcement learning.
In principle, there are several ways to extract such a latent policy prior from a pretrained LLM.
Since the main focus of this work is on how the prior is incorporated into online reinforcement learning, we adopt a KALM-style approach in this paper to obtain the prior.

Specifically, given an offline expert dataset, we first provide the task description as textual input to the pretrained LLM.
We then learn a projection module to convert the non-textual state-action sequence from offline demonstrations into latent representations and feed them into the LLM together with the task description.
Conditioned on both the task description and the projected demonstration trajectory, a newly added action head is trained to predict the next action.

Formally, let $x$ denote the task description, let $\tau_t=\{(s_1,a_1), (s_2,a_2),\cdots,(s_t,a_t)\}$ denote the state-action trajectory prefix from an offline demonstration, let $E_s(\cdot)$ and $E_a(\cdot)$ denote the learnable projection modules for states and actions, let $F_\phi(\cdot)$ denote the pretrained LLM backbone, and let $G_\theta(\cdot)$ denote the newly added action head.
At time step $t$, each state and action in $\tau_t$ is first mapped into latent embeddings as
\begin{equation}
e_i^s = E_s(s_i), \qquad e_i^a = E_a(a_i), \quad i=1,\dots,t.
\end{equation}
These embeddings, together with the task description, are then fed into the LLM as a sequence
\begin{equation}
\mathcal{X}_t = [x,e_1^s,e_1^a,e_2^s,e_2^a,\dots,e_t^s,e_t^a].
\end{equation}
The resulting latent representation is computed as
\begin{equation}
z_t = F_\phi(\mathcal{X}_t),
\end{equation}
and the latent policy prior is defined as
\begin{equation}
\pi_K(a_{t+1}\mid x,\tau_t) = G_\theta(z_t).
\end{equation}

We train this latent policy prior using an offline expert dataset $\mathcal{D}=\{(x,\tau_t,a_{t+1})\}$.
The training objective is to minimize the negative log-likelihood:
\begin{equation}
    \mathcal{L}_{\text{offline}} = -\mathbb{E}_{(x,\tau_t,a_{t+1})\sim \mathcal{D}} \left[ \log \pi_K(a_{t+1}\mid x,\tau_t) \right].
    \label{eq:kalm_loss}
\end{equation}


Overall, this stage is not intended to produce an optimal LLM controller, as it uses supervised fine-tuning rather than reward optimization.
Instead, this model aims to capture a coarse behavioral prior from offline demonstrations and pretrained LLM knowledge, which will later be used to guide online RL policy learning.

\subsection{Online Reinforcement Learning with Latent Action Guidance}
\label{subsec:lago_stage2}

After learning the latent policy prior $\pi_K$ from offline demonstrations, we use it to guide online reinforcement learning in the second stage.
Specifically, we freeze the prior model and train a separate RL policy $\pi_\psi(a|s)$ through online interaction with the environment.
Instead of learning only from task rewards, LaGO regularizes the policy learning toward the latent policy prior produced by the pretrained LLM.
The training loss is defined as:
\begin{equation}
    \mathcal{L}_{\text{online}}        
        = \mathcal{L}_{\text{PG}} + \beta\, \mathcal{L}_{\text{prior}} \\
\label{eq:rl_loss}
\end{equation}
where $\mathcal{L}_{\text{PG}}$ denotes a standard policy optimization objective, such as PPO or SAC, and $\beta$ controls the influence of the prior.

In practice, the prior regularization term is implemented differently for discrete and continuous action spaces.
The loss is defined as:
\begin{equation}
    \mathcal{L}_{\text{prior}}
    =
    \begin{cases}
    -\log \pi_\psi(a_K\mid s),
    & \text{discrete}, \\[4pt]
    -\displaystyle\int_{\mathcal{A}}
    \pi_K(a\mid s)\log \pi_\psi(a\mid s)\,da,
    & \text{continuous},
    \end{cases}
    \label{eq:kl_surrogate}
\end{equation}
where $a_K$ denotes the action predicted by the latent policy prior, and $\mathcal{A}$ denotes the continuous action space.
The continuous cross-entropy term differs from the exact KL only by the entropy of the fixed prior distribution, and therefore provides an equivalent optimization signal for policy regularization.
Thus, the latent prior acts as a soft behavioral bias that guides policy learning while still allowing the RL policy to adapt to online reward feedback.

The training procedure is summarized in Algorithm~\ref{alg:training-procedure}.
Overall, this design allows LaGO to benefit from pretrained LLM knowledge even when the latent prior is not sufficiently accurate to serve as a standalone controller.
Instead, the prior only needs to provide a coarse but informative behavioral bias, which can guide policy learning toward more promising directions during training.
This is particularly useful in challenging tasks, where reward signals are limited and unguided exploration can be inefficient.
Moreover, since the influence of the prior is controlled by the coefficient $\beta$, the guidance remains soft and does not overly constrain policy learning when the prior is imperfect.
By combining such rough prior guidance with online environment feedback, the RL policy remains adaptive while still benefiting from the structural knowledge encoded in the pretrained LLM.

\begin{algorithm}[t]
\caption{LaGO Training Procedure}
\label{alg:training-procedure}
\begin{algorithmic}[1] 
\STATE \textbf{Input:} Offline dataset $\mathcal{D}$, pretrained LLM backbone $F_\phi$, regularization coefficient $\beta$
\STATE \textbf{Stage 1: Offline Prior Learning}
\STATE Train projection layers and action head on $\mathcal{D}$ to minimize $\mathcal{L}_{\text{offline}}$
\STATE Obtain latent policy prior $\pi_K$
\STATE \textbf{Stage 2: Online RL}
\STATE Freeze latent policy prior $\pi_K$
\STATE Initialize RL policy $\pi_\psi(a \mid s)$
\REPEAT
    \STATE Collect rollouts via interaction with the environment
    \STATE Compute policy objective $\mathcal{L}_{\text{PG}}$
    \STATE Compute prior loss $\mathcal{L}_{\text{prior}}$
    \STATE Update policy parameters $\psi$ using $\mathcal{L}_{\text{online}} = \mathcal{L}_{\text{PG}} + \beta\,\mathcal{L}_{\text{prior}}$
\UNTIL{Convergence}
\STATE \textbf{Output:} Optimized policy $\pi_\psi(a \mid s)$
\end{algorithmic}
\end{algorithm}
\begin{table*}[th]
\caption{%
    Performance comparison on CLEVR-Robot and Meta-World tasks (Last 10 steps Mean $\pm$ Std).
    We report reward and success rate for Vanilla PPO and LaGO.
    For Vanilla PPO, the \textit{Rephrase} baseline reuses the corresponding \textit{Real} task result because the policy is instruction-agnostic.
    Best results are highlighted in \textbf{bold}.
}
\label{tab:main_results}
\vspace{2pt}
\centering
{\footnotesize
\setlength{\tabcolsep}{4pt}
\renewcommand{\arraystretch}{0.97}
\begin{tabular}{llcccc}
\toprule
\textbf{Task} & \textbf{Method} & \multicolumn{2}{c}{\textbf{CLEVR-Robot}} & \multicolumn{2}{c}{\textbf{Meta-World}} \\
\midrule
\multicolumn{2}{c}{} & \textbf{Reward} & \textbf{Success Rate} & \textbf{Reward} & \textbf{Success Rate} \\
\midrule
\textit{Real} & Vanilla PPO & $0.030 \pm 0.004$          & $0.230 \pm 0.022$          & $0.741 \pm 0.077$          & $0.023 \pm 0.026$          \\
              & LaGO        & $\mathbf{0.078 \pm 0.003}$ & $\mathbf{0.413 \pm 0.037}$ & $\mathbf{1.167 \pm 0.023}$ & $\mathbf{0.233 \pm 0.022}$ \\
\midrule
\textit{Rephrase} & Vanilla PPO & $0.030 \pm 0.004$          & $0.230 \pm 0.022$          & $0.741 \pm 0.077$          & $0.023 \pm 0.026$          \\
                  & LaGO        & $\mathbf{0.069 \pm 0.002}$ & $\mathbf{0.388 \pm 0.041}$ & $\mathbf{1.130 \pm 0.105}$ & $\mathbf{0.235 \pm 0.026}$ \\
\midrule
\textit{Easy} & Vanilla PPO & $0.138 \pm 0.021$          & $0.138 \pm 0.021$          & $1.108 \pm 0.216$          & $0.057 \pm 0.035$          \\
              & LaGO        & $\mathbf{0.148 \pm 0.014}$ & $\mathbf{0.148 \pm 0.014}$ & $\mathbf{1.343 \pm 0.038}$ & $\mathbf{0.082 \pm 0.002}$ \\
\midrule
\textit{Hard} & Vanilla PPO & $0.106 \pm 0.007$          & $0.007 \pm 0.005$          & $0.768 \pm 0.015$          & $0.005 \pm 0.006$          \\
              & LaGO        & $\mathbf{0.194 \pm 0.017}$ & $\mathbf{0.140 \pm 0.016}$ & $\mathbf{1.002 \pm 0.011}$ & $\mathbf{0.058 \pm 0.018}$ \\
\midrule
\textit{Overall Average} & Vanilla PPO & $0.076$                    & $0.151$                    & $0.840$                    & $0.027$                    \\
                         & LaGO        & $\mathbf{0.122}$           & $\mathbf{0.272}$           & $\mathbf{1.161}$           & $\mathbf{0.152}$           \\
                         & Improvement ($\Delta$) & $\mathbf{+0.046}$ & $\mathbf{+0.121}$ & $\mathbf{+0.321}$ & $\mathbf{+0.125}$ \\
\bottomrule
\end{tabular}
}
\end{table*}

\section{Experiments}
\label{sec:experiments}


\subsection{Experimental Setup}
\label{subsec:experimental_setup}

We evaluate LaGO on two control benchmarks: CLEVR-Robot~\cite{google_research2019clevr_robot_env}, which has a discrete action space, and Meta-World~\cite{yu2020meta}, which has a continuous action space.
CLEVR-Robot is a manipulation benchmark in which the agent moves five balls to satisfy target spatial relations.
Its offline dataset covers tasks that move a designated ball relative to another ball in directions such as front, behind, left, and right.
The state is represented by a 10-dimensional vector of ball positions, and the action is a 40-dimensional one-hot vector representing one of 40 discrete movements.
Meta-World is a robotic manipulation benchmark in which the agent controls a Sawyer robot to interact with objects such as doors, drawers, windows, etc.
Its offline dataset includes a diverse set of manipulation tasks, such as reach, push, pick-place, button-press, door-related tasks, window-open, faucet-open, and coffee-related tasks.
The state is represented by a 91-dimensional vector describing the robot state and object poses, and the action is a 4-dimensional gripper control signal.

For each benchmark, we follow the settings used in KALM~\cite{pang2024knowledgeableagentsofflinereinforcement} and consider four task categories: \textit{Real}, \textit{Rephrase}, \textit{Easy}, and \textit{Hard}.
Concretely, \textit{Real} denotes tasks already covered by the offline data.
\textit{Rephrase} keeps the same underlying task but replaces the instruction with a paraphrased natural-language description.
\textit{Easy}/\textit{Hard} denote unseen tasks outside the offline dataset, with Hard increasing novelty and compositional difficulty.
We report both reward and success rate as the evaluation metrics.

For the main experiments, we use Llama-2-7b-chat-hf as the language backbone and PPO as the Stage 2 online RL optimizer.
For Stage 1 latent policy pretraining, we follow the grounding schedule of KALM and train for 10 epochs on CLEVR-Robot and 5 epochs on Meta-World.
For Stage 2 online RL, we train for 1.5M environment steps on CLEVR-Robot and 20M environment steps on Meta-World.
All experiments are conducted on a machine with four NVIDIA RTX A6000 GPUs.

\subsection{Main Results}
\label{subsec:main_results}

Table~\ref{tab:main_results} summarizes the main results on CLEVR-Robot and Meta-World.
Overall, LaGO consistently outperforms Vanilla PPO on both benchmarks in terms of both reward and success rate, showing that the latent policy prior provides a useful learning signal for online reinforcement learning in both discrete and continuous control settings.
On CLEVR-Robot, LaGO improves the overall average reward from 0.076 to 0.122 and the overall average success rate from 0.151 to 0.272.
On Meta-World, LaGO improves the overall average reward from 0.840 to 1.161 and the overall average success rate from 0.027 to 0.152.
Moreover, these improvements are consistently observed across all four task categories.

The improvements are most evident on the \textit{Real} and \textit{Rephrase} categories in both benchmarks.
We believe this is because these tasks are more closely aligned with the offline data used to learn the latent policy prior in Stage 1.
As a result, the LLM-based prior can provide more reliable task-relevant guidance on these in-distribution settings.
However, LaGO still improves performance on the unseen \textit{Easy} and \textit{Hard} categories.
This suggests that the prior does not merely memorize the seen tasks but also provides a certain degree of generalization ability from the pretrained language model.
Even when the target task is not covered during training, the prior can still provide a rough but useful policy bias, which helps guide the online RL policy toward better directions.

In addition, from the success rates in both Vanilla PPO and LaGO, Meta-World is clearly more challenging than CLEVR-Robot.
This is because Meta-World involves continuous action control, whereas CLEVR-Robot has a discrete action space.
Nevertheless, the overall gain in success rate remains remarkably similar across the two benchmarks: +0.121 on CLEVR-Robot and +0.125 on Meta-World.
This suggests that the latent policy prior can still provide a useful training signal even in the more challenging continuous-control setting.

\begin{figure}[h]
    \centering
    \includegraphics[width=\linewidth]{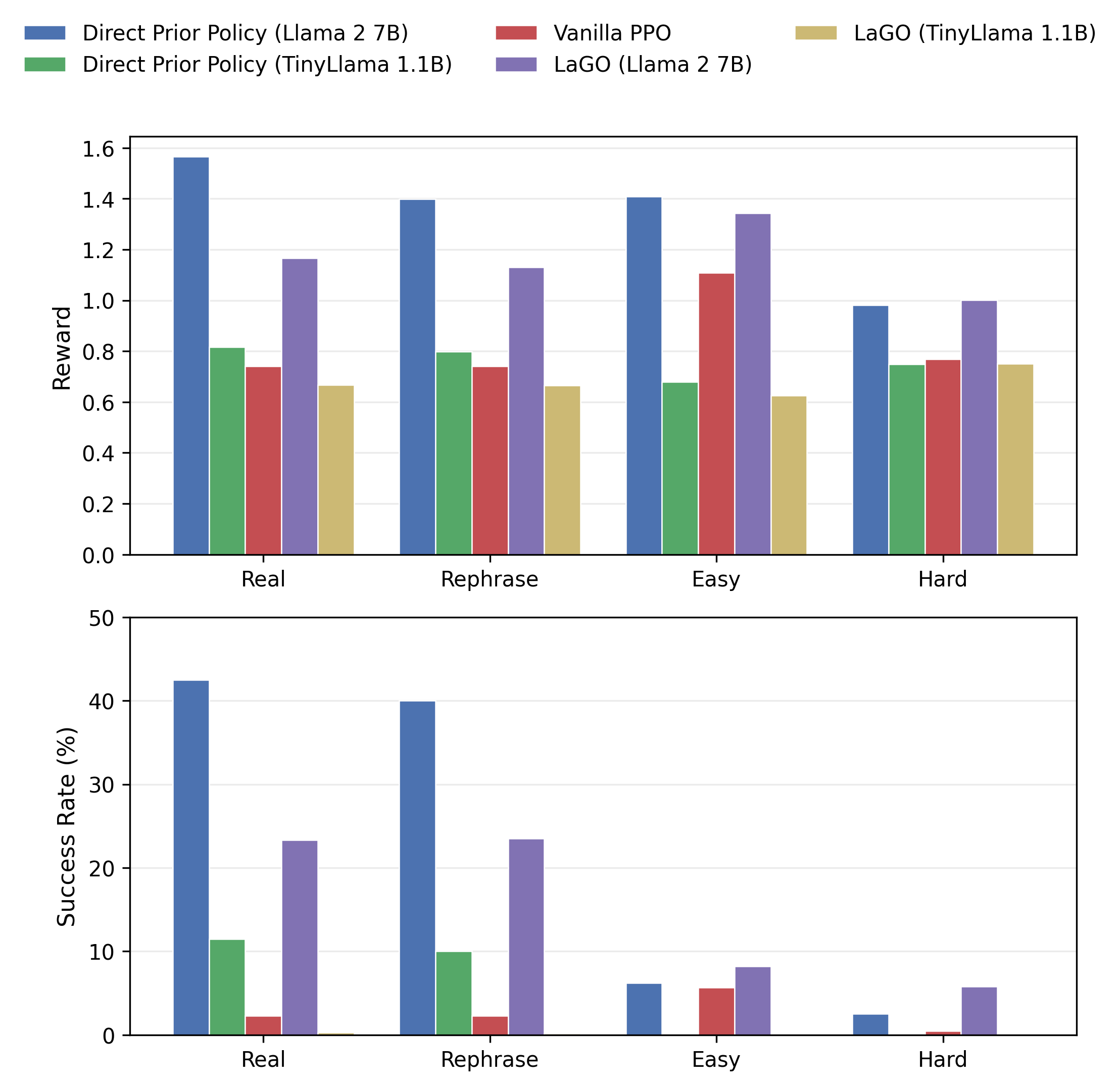}
    \caption{%
        Impact of LLM prior quality on Meta-World.
        For each task category, we report reward and success rate for the direct prior policy, Vanilla PPO, and LaGO.
    }
    \label{fig:metaworld_backbone_bars}
\end{figure}

\subsection{Impact of LLM Prior Quality}
\label{subsec:backbone_ablation}

In this subsection, we examine the effect of LLM prior quality on the performance of LaGO.
Specifically, we train two LLM priors based on Llama 2 7B and TinyLlama 1.1B within the same LaGO framework on Meta-World.
This comparison allows us to assess whether a stronger pretrained language model leads to a more useful latent policy prior for online RL training.

Figure~\ref{fig:metaworld_backbone_bars} shows the reward and success rate across task categories for three types of methods: Direct Prior Policy, Vanilla PPO, and LaGO with different LLM priors.
Here, Direct Prior Policy means that the learned prior policy is directly executed in the environment without Stage 2 online RL training.
Namely, after Stage 1 training, the prior policy is used to interact with the environment directly.
A clear trend can be observed in the Direct Prior Policy.
The prior learned from Llama 2 7B consistently outperforms the prior learned from TinyLlama 1.1B in both reward and success rate across all categories.
This suggests that a stronger pretrained language model preserves more task-relevant information after Stage 1 training.
Note that although the Direct Prior Policy performs better than PPO and LaGO, the model sizes are drastically different.
The LLM backbone contains 1.1B or 7B parameters, whereas the PPO policy used in Stage 2 contains only approximately 57 thousand parameters.
Thus, the direct prior achieves strong performance, but it relies on more parameters than PPO.

Next, a similar trend is also shown in the online RL results.
LaGO with Llama 2 7B consistently achieves the strongest overall performance among the learned policy variants.
In contrast, using the prior trained from TinyLlama 1.1B leads to clear drops in both reward and success rate, and in some cases even performs worse than Vanilla PPO.
This indicates that the quality of the pretrained language model has a substantial impact on the effectiveness of the learned prior.

Overall, these results suggest that LaGO can effectively benefit from improvements in LLM prior quality.
As the pretrained language model becomes stronger, the resulting latent prior also becomes more useful for guiding online RL training.
This highlights an important advantage of LaGO: future advances in pretrained language models can be directly translated into stronger latent guidance within our framework.


\section{Discussion}
\label{sec:discussion}

This paper investigates whether pretrained LLMs can improve decision-making in planning problems, not by acting as standalone controllers, but by serving as latent action priors that softly guide policy optimization in online reinforcement learning.
Across both CLEVR-Robot and Meta-World, LaGO consistently improves both reward and success rate over Vanilla PPO.
These results suggest that LLMs' knowledge can provide a useful inductive bias for subsequent online learning.
Moreover, using a stronger foundation model leads to a more effective latent prior, suggesting that the quality of the pretrained LLM is an important factor in determining the benefit of such guidance.
Overall, our findings support a simple but important conclusion: the knowledge and generalization ability encoded in pretrained LLMs can serve as a strong prior for planning and reinforcement learning.

Several directions remain for future work.
First, since our results show that stronger pretrained LLMs lead to better guidance, a straightforward next step is to study LaGO with more capable foundation models.
Second, the current framework applies prior guidance throughout training, while a more adaptive design that gradually reduces the weight of the prior as the RL policy becomes stronger may further improve performance.
Third, it would be valuable to evaluate LaGO in broader and more challenging settings, such as longer-horizon tasks and more complex decision-making problems.
More broadly, we believe this direction opens up a practical way to leverage pretrained language models for reinforcement learning without requiring them to generate precise actions, plans, or trajectories during interaction.




\section*{Impact Statement}
This paper presents work whose goal is to advance the field of Machine Learning.
There are many potential societal consequences of our work, none of which we feel must be specifically highlighted here.

\section*{Acknowledgement}
This research is partially supported by the National Science and Technology Council (NSTC) of the Republic of China (Taiwan) under Grant Number NSTC 113-2221-E-001-009-MY3, NSTC 114-2634-F-A49-004, NSTC 114-2221-E-A49-005. The authors would also like to thank the anonymous reviewers for their valuable comments.

\bibliography{references}
\bibliographystyle{icml2026}


\end{document}